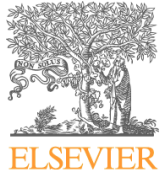

# A promotion method for generation error-based video anomaly detection

Zhiguo Wang[a], Zhongliang Yang[a] and Yujin Zhang [a], ∗

[a] *Image Engineering Laboratory, Tsinghua University, Beijing, 100084, China*

ABSTRACT

Surveillance video anomaly detection is to detect events that rarely or never happened in a certain scene. The generation error (GE)-based methods exhibit excellent performance on this task. They firstly train a generative neural network (GNN) to generate normal samples, then judge the samples with large GEs as anomalies. Almost all the GE-based methods utilize frame-level GEs to detect anomalies. However, anomalies generally occur in local areas, the frame-level GE introduces GEs of normal areas to anomaly discriminations, that brings two problems: i) The GE of normal areas reduces the anomaly saliency of the anomalous frame. ii) Different videos have different normal-GE-levels, thus it is hard to set a uniform threshold for all videos to detect anomalies. To address these problems, we propose a promotion method: utilize the maximum of block-level GEs on the frame to detect anomaly. Firstly, we calculate the block-level GEs at each position on the frame. Then, we utilize the maximum of the block-level GEs on the frame to detect anomalies. Based on the existed GNN models, experiments are carried out on multiple datasets. The results demonstrate the effectiveness of the proposed method and achieve state-of-the-art performance.

## 1. Introduction

Surveillance videos are important for social security. However, it is time-consuming and labor-intensive to watch surveillance videos for a long time. Therefore, it is necessary to detect video anomalies automatically. It is challenging, because the anomalies rarely happen and the types of anomalies are uncountable.

The development of video anomaly detection can be divided into two stages: traditional machine learning stage [1,2] and deep learning stage [3]. In the deep learning stage, using the GE of a GNN to detect anomalies is an important branch. They firstly train a GNN to generate (reconstruct or predict) normal samples, then judge the samples with large GEs as anomalies.

Utilizing the trained GNN model and the ground truth of the output of GNN, we can calculate the GE (intensity) map. The GE maps have two characters: i) The GE mainly existed on the foreground targets area. The reason is that the background is static, the information of the background is easy to learn, while the foreground targets change a lot over time. ii) The GE of the anomalous areas are larger than the GE of the normal areas. The reason is that the GNNs have learned the knowledge about normal events, but not about abnormal events. An example of GE map is shown in Fig. 1.

In the testing period, after the GE maps been generated, most of the existed works utilize the frame-level GE to detect anomalies, which has a shortcoming: The GE of normal areas disturbs anomaly detection. i) They reduce the anomaly saliencies of the anomalous frames. The reasons are as follows. In anomalous frame, the frame-level GE averages the GE of anomalous area with the GE of normal area, that reduces the impact of the GE of anomalous area to the final GE which is used to detect anomalies, thus reduces anomaly saliency. ii) They result in that different video segments have different normal GE-levels, which causes that it is hard to set a unified threshold for all video segments to detect anomalies. The reason for this problem is that different videos have different numbers of foreground targets, while GE exists in every foreground target area. Many works [4][5][6] solve this problem by normalizing the GEs in each video segment. However, this method brings another problem: it produces high anomaly scores in each video segment, even if there is no anomaly existed. This is not in line with the application needs of our real life.

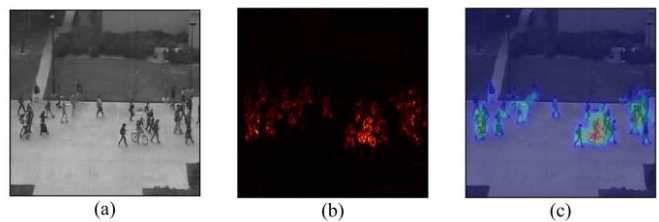

**Fig. 1.** The GE map generated by the GNN in work [6]. (a) The ground truth of the output. (b) The GE intensity map. (c) The GE heat map.

In this paper, we propose a promotion method to solve the above problems. We reduce the disturbance of the GE in normal areas to anomaly detection. In the testing period, after the GE maps been generated, we firstly calculate the block-level GE at each position on the scene. This process can be implemented by a mean filter operation. Then, we utilize the maximum of the block-level GEs in the frame to detect anomalies. The pipeline of our method is illustrated in Fig. 2.

───────────

∗ Corresponding author. e-mail: zhang-yj@tsinghua.edu.cn

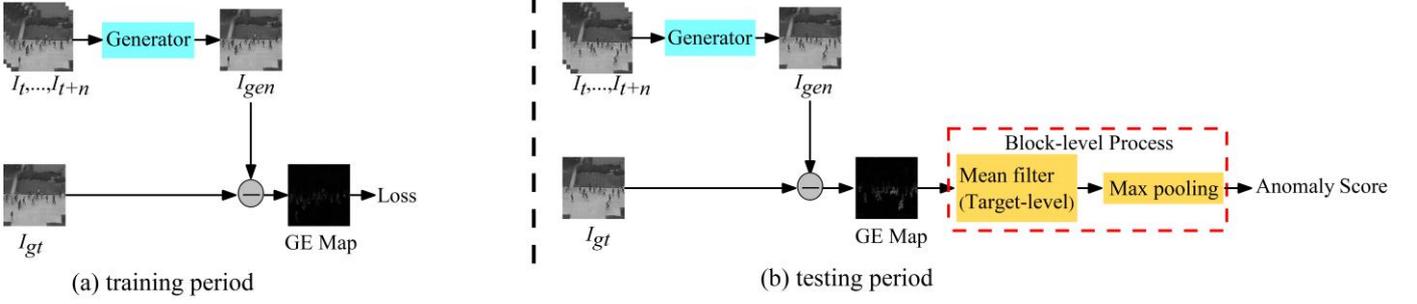

**Fig. 2.** The pipeline of the proposed method. (a) Training period. We train a generator to generate normal samples. In this process, we use frame-level GE as loss to learn the information of the whole scene. (b) Testing period. After the GE map for a test frame been calculated, we utilize a block-level process module to calculate a block-level GE for the frame, and then use the block-level GE to calculate anomaly score and to detect anomaly.

We summarize our contributions as follows: We propose a promotion method for GE-based methods to reduce the interference of the GEs in normal areas to anomaly detection, thus to improve their anomaly detection performance. We carry out experiments on multiple datasets. The results prove the effectiveness of the proposed method and achieve state-of-the-art performance on multiple datasets.

## 2. Related work

The development of video anomaly detection can be divided into two stages: traditional machine learning stage [1,2] and deep learning stage [3][7].

The algorithms in the traditional machine learning stage usually use hand-crafted-features [8–20] or deep-features [21–28] to construct feature space, and then utilize the traditional machine learning models to detect the outliers. Mahadevan et al.[17] utilized the mixture of dynamic textures (MDT) to detect the temporal anomalies. Kratz et al.[15] utilized hidden Markov model (HMM) to learn the natural motion transitions in each local area, and coupled multiple HMMs to model the spatial relationship between nearby regions. Hu et al. [13] utilized the topic model to detect anomalies. They divided the normal features into different topics, and classified the samples which do not belong to any existed topics as anomalies. Hinami et al. [22] and Tran et al.[27] utilized pretrained convolutional neural network (CNN) to extract features and used one-class support vector machine (OC-SVM) to detect anomalies. After the anomalies detected, Hinami et al. [22] used the semantic information in the features to recount the anomalies. Besides the above methods, sparse coding (SC) is also a popular method to detect anomalies [8–10,12,16,18,20,21,26]. They firstly learn an over-complete dictionary to reconstruct normal samples with small reconstruction errors, and then judge the samples with large reconstruction errors as anomalies. The process of SC is time-consuming. To accelerate the process, Lu et al. [16] proposed to learn multiple small dictionaries to encode the normal samples.

The algorithms in the deep learning stage can be divided into three categories:

i) Cluster-based methods [29][30][31]. They firstly utilize neural networks to classify samples into multiple clusters, and then determine the samples do not belong to any existed clusters as anomalies. Fan et al. [31] utilized gaussian mixture variational autoencoder (VAE) to classify samples into multiple clusters, then utilized the conditional probabilities of the test samples to detect anomalies. The samples with low conditional probabilities on all clusters were judged as anomalies.

ii) Generative adversarial network (GAN)-based methods [32][33][34]. They firstly utilize GAN to learn the manifold distribution of normal samples by generating normal samples. Then, they utilize the discriminator to detect anomalies. Tang et al. [32] utilized U-net to predict and reconstruct the future frames, then utilized discriminator to detect anomalies.

iii) GE-based methods [4][6][33][35][36]. They firstly utilize GNN to learn the manifold distribution of normal samples by reconstructing the input frames [4][35], predicting the future frames [6], generating another modality of the inputs [33][36], or generating the combination of the above [32][37][38,39]. Then they utilize the GE to measure the deviation of the test samples from the normal distribution. The commonly used GNNs include: autoencoder (AE) [4,40], long-short-term-memory AE (LSTM-AE) [5,35,38,39], U-net [6,32][36], sparse AE [41], VAE [42]. In the testing phase, they usually utilize frame-level GE to detect anomalies. For example, Hasan et al. [4] utilized AE as GNN, and utilize the frame-level GEs to detect anomalies.

In some cases, the anomaly scores generated by GE maps and that generated by discriminator are combined [5][43]. For example, Lee et al. [5] utilized a bidirectional LSTM-AE as GNN to generate the inter-frame in a sequence, and used a 3D convolutional discriminator to determine whether the generated sequence is real. Finally, they combined the anomaly scores generated by GE maps and that generated by discriminator to detect anomalies.

In this paper, we focus on the GE-based methods.

## 3. Our method

The process of GE-based deep learning methods can be divided into two steps: training step and testing step. Our promotion method is utilized in the testing step.

### 3.1. GE-maps

After the GNN been trained, we firstly utilize GNN to generate GE maps. Let $\mathcal{G}$ be the GNN, $\hat{I}$ be the output of $\mathcal{G}$, $I$ be the ground truth of $\hat{I}$. We calculate the GE map $E$ with the following formula:

$$E_{i,j} = \sum_{c=1}^{C} \| \hat{I}_{i,j} - I_{i,j} \|_{\mathcal{L}_n}, \quad (1)$$

where $C$ is the channel number of $I$, $i$ and $j$ are spatial coordinates on the frame, $\mathcal{L}_n$ means the norm when calculating the errors.

### 3.2. Block-level process

In a video segment, the GE saliency of an abnormal frame compared to a normal frame, which we term as anomaly saliency, can be written as:

$$Saliency = \frac{L_{abnormal} - L_{normal}}{L_{normal}}, \quad (2)$$

$$L_{abnormal} = \frac{\sum_{j=1}^{n} e'_j + \sum_{i=1}^{P-n} e_i}{P}, \quad (3)$$

$$L_{normal} = \frac{\sum_{i=1}^{n} e_i + \sum_{i=n}^{P} e_i}{P}, \quad (4)$$

where $L_{abnormal}$ and $L_{normal}$ are GEs of an abnormal frame and a normal frame respectively, $P$ is the number of pixels on $I$, $e_i$ is the GE value of the $i$-th normal pixel on the normal or abnormal GE map, $e'_j$ is the GE value of the $j$-th anomaly pixel on the abnormal frame, $n$ is the number of abnormal pixels on the abnormal frame.

We can assume that the GE of the normal area in an abnormal frame is equal to the GE of the corresponding normal area in a normal frame. We substitute formula (3)(4) into formula (2), then the anomaly saliency can be written as:

$$Saliency = \frac{\sum_{j=1}^{n} e'_j - \sum_{i=1}^{n} e_i}{\sum_{i=1}^{n} e_i + \sum_{i=n}^{P} e_i}. \quad (5)$$

From equation (5), we can find that the anomaly saliency is negatively correlated to $\sum_{i=n}^{P} e_i$, which means the GE of normal areas in the anomalous frame. That indicate that the more the GE of normal regions contributes to the final GE, the lower the GE saliency of the abnormal frame.

Therefore, by reducing the contribution of the GE of normal regions to the final GE, we can improve anomaly saliencies. In this paper, we achieve this goal by substituting the frame-level GE with block-level GE: We firstly put a size-fixed sliding window on the frame, and calculate the block-level GE at each window position. Then, we select the maximum of the block-level GEs on the frame to detect anomalies.

$$L_{B_k}(t) = \frac{1}{h*w} \sum_{i=1}^{h} \sum_{j=1}^{w} B_{k,i,j}, \quad (6)$$

$$L_B(t) = max\{L_{B_1}(t), ..., L_{B_K}(t)\}, \quad (7)$$

where $L_{B_k}(t)$ means the block-level GE in the $k$-th block $B_k$ on the frame, $K$ is the total number of the blocks on a frame, $h$ and $w$ are the height and width of the block, respectively, $L_B(t)$ is the maximum of all the $L_{B_k}(t)$ on a frame. Note that the block-level operation can be implemented by a convolution layer for mean filter and a max-pooling layer. Therefore, this operation can be accelerated on the GPU. After the block-level process, we use a median filter to smooth the GEs along the time axis. In this paper, we set the median filter radius as 15.

### 3.3. Anomaly Score

When utilizing the GE values to calculate the anomaly scores, most of the existed GE-based methods [4][5][6] utilize the max-min normalization strategy to normalize the GEs in each video. Their reason is that different video segments have different normal-GE-levels.

In this paper, we calculate anomaly scores by normalizing GEs in the whole dataset instead of in each video. Our reasons are as follows: i) The block-level process improves the anomaly saliencies, that alleviates the problem brought by different normal GE-levels. ii) The different normal-GE-levels are mainly caused by different foreground target numbers in the scene. Block-level process reduces the relativity between the foreground target numbers and the normal-GE-levels, and thus reduces the difference between different normal-GE-levels. iii) Normalizing the GEs in each video segment produces high anomaly scores in each video segment, which results in that anomalies would be detected in each video segments even if there is no anomaly existed. This makes the algorithms inapplicable to the situations where a large number of normal samples are existed.

Therefore, we utilize equation (8) to calculate the anomaly scores.

$$s(t) = \frac{L_B(t) - min(L_B)}{max(L_B) - min(L_B)}, \quad (8)$$

where the $min(L_B)$ and $max(L_B)$ are, respectively, the minimum and maximum values of the GEs of all the frames in the dataset.

When multiple anomaly scores are available for a frame, we utilize the weighted sum of them to calculate the final anomaly score:

$$\hat{s}(t) = \sum_{i=1}^{n} \alpha_i s_i(t), \quad (9)$$

where $s_i(t)$ is the $i$-th anomaly score for the $t$-th frame in dataset, $\alpha_i$ is the weight for $s_i(t)$.

## 4. Experiments

In this section, we utilize work [6] as baseline to evaluate our method. The reason for choosing work [6] as baseline are as follows: i) Work [6] is a typical GE-based method. ii) Multiple GE maps are available in this work, include: the GE map of the pixel value $GE_{pixel}$, the GE map of the optical flow $GE_{flow}$, and the GE map of the gradient value $GE_{gdl}$, that is convenient for us to evaluate our method more sufficiently.

We evaluate the effectiveness of the promotion method from three perspectives: anomaly saliency, normal GE-level, anomaly detection performance. Then we analyze the impact of block-size and different normalization strategies to anomaly detection. Finally, we compare the proposed method with the existed methods.

### 4.1. Dataset and Evaluation Criteria

The experiments are carried out on two datasets: CUHK Avenue dataset [16] and UCSD Pedestrian dataset [17]. The Avenue dataset contains 16 training videos and 21 testing videos. The abnormal events include running, throwing schoolbag, throwing papers, etc. The UCSD dataset has two sub-datasets: Ped1, Ped2. The two sub-datasets capture different scenarios but have the similar definition of abnormal events, include cycling, skateboarding, crossing lawns, cars, etc. These two sub-datasets are usually used separately.

The most commonly used evaluation metric is the Receiver Operation Characteristic (ROC) and the Area Under Curve (AUC). The higher the AUC, the better the anomaly detection performance. Following the work [6], we detect the frame-level anomalies and use the frame-level AUC for performance evaluation.

### 4.2. Impact of block-level process to anomaly saliency

In this section, we firstly visualize the frame-level GE curve and block-level GE curve. Then, we calculate and compare the anomaly saliencies before and after block-level process on multiple datasets and multiple types of GE maps.

The frame-level GE curve and block-level GE curve are shown in Fig. 3. In the block-level GE curve, $h = w = 30$.

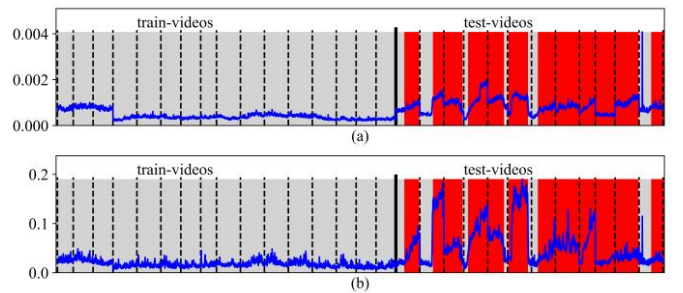

**Fig. 3.** The GE curves of $GE_{pixel}$ in Ped2. The horizontal axis indicates the indexes of the frames. The vertical axis indicates the GEs of different frames. Vertical dashed lines separate different video samples. The frames in the gray areas are normal frames, and the frames in the red areas are abnormal frames. (a) The frame-level GE curve. (b) The block-level GE curve.

As shown in Fig. 3, in the block-level GE curve, the GE of the anomalous frames are more salient compared with that in the frame-level GE curve, which is better for anomaly detection task.

The anomaly saliencies before and after block-level process in multiple datasets and multiple types of GE maps are shown in Table 1, where we use the average of the GEs of all the normal frames to replace $L_{normal}$, and use all the average of the GEs of abnormal frames to replace $L_{abnormal}$.

**Table 1**
Anomaly saliences in multiple datasets and multiple generation error maps. The higher the anomaly saliency, the better.

| Generation Error | | Ped1 | Ped2 | Avenue |
|---|---|---|---|---|
| $GE_{pixel}$ | Frame-level | 1.158 | 0.7475 | 2.6341 |
| | **Block-level** | **2.5239** | **2.4162** | **3.4552** |
| $GE_{flow}$ | Frame-level | 1.0667 | 0.7212 | 2.4865 |
| | **Block-level** | **1.6224** | **1.4112** | **2.7777** |
| $GE_{gdl}$ | Frame-level | 0.1119 | 0.1141 | 0.2965 |
| | **Block-level** | **0.4454** | **0.6714** | **0.7185** |

As shown in Table 1, the anomaly saliencies in block-level GEs are much higher than that in the frame-level GEs, which proves that the block-level process can help to improve the anomaly saliencies.

*4.3. Impact of block-level process to normal GE-levels*

Different video segments have different normal GE-levels, that disturbs anomaly detection. In this section, we prove the effectiveness of block-level process to solve this problem from three aspects: i) Prove that normal GE-levels are positively correlated with foreground targets numbers. ii) Prove that the block-level process can reduce the correlation between the foreground targets numbers and the normal GE-levels. iii) Prove that the block-level process can reduce the difference between different normal GE-levels of different video segments.

Fig. 4 visualizes the relationship between normal GE-levels and foreground targets numbers in Ped2 dataset. The GEs are calculated based on $GE_{pixel}$, the parameters in block-level process are: $h = w = 30$. In Ped2 dataset, all the video segments are short, we can assume that the number of objects in a video segment is steady. Therefore, we calculate the average of objects numbers in each video. In order to visualize the GE curve and foreground targets numbers curve in a same figure, we divide the foreground targets numbers by a fixed value.

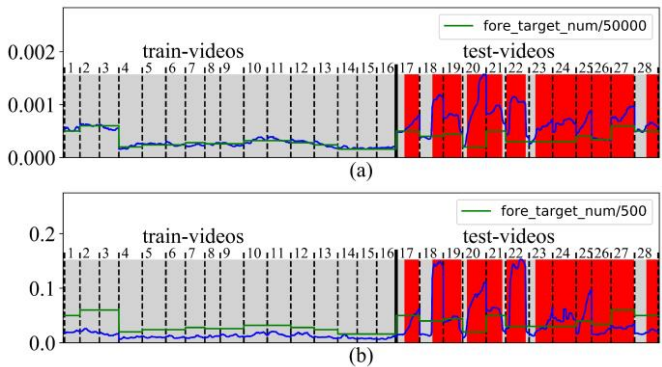

**Fig. 4.** Visualization of the relationship between normal GE-levels and foreground targets numbers in Ped2 dataset. The blue curve indicates the GE of different frames. The green curve reflects the averages of foreground targets numbers in different video segments. The gray area indicates normal frames, the red area indicates abnormal frames. (a) The relationship between frame-level GEs and foreground targets numbers. (b) The relationship between block-level GEs and foreground targets numbers.

Table 2 shows the correlation coefficients in Ped2 dataset in multiple types of GE maps.

**Table 2**
The correlation coefficients between the number of foreground targets and the normal GE-levels. The lower the correlation coefficients, the better.

| Generation Error | | Ped2 |
|---|---|---|
| $GE_{pixel}$ | Frame-level | 0.9843 |
| | **Block-level** | **0.8528** |
| $GE_{flow}$ | Frame-level | 0.9818 |
| | **Block-level** | **0.8822** |
| $GE_{gdl}$ | Frame-level | 0.985 |
| | **Block-level** | **0.8461** |

From Fig. 4 and Table 2, we can find that the correlation between foreground target numbers and normal GE levels are high, and block-level process can reduce this correlation.

In Table 3, we utilize several couples of samples in Ped2 to show the effectiveness of the block-level process to reduce the difference between different normal-GE-levels. We utilize the ratios of different normal-GE-levels to measure their difference. We utilize the average of GEs of normal frames in a video segment to calculate the normal-GE-level, and divide the higher normal-GE-level by lower normal-GE-level to calculate ratios to evaluate their difference. The closer the ratio gets to 1, the better. The video indexes are same to that shows in Fig. 4. As shown in Table 3, the block-level process can reduce the difference between different normal GE-levels.

**Table 3**
Ratios of different normal GE-levels of different video segments. We calculate the ratios by dividing the higher normal GE-level by the lower GE-level. The closer the ratio gets to 1, the better.

| Video index | Foreground target number | Ratio in frame-level GEs | Ratio in block-level GEs |
|---|---|---|---|
| 3 | 30 | 2.6305 | **1.6264** |
| 4 | 10 | | |
| 17 | 25 | 2.0822 | **1.2875** |
| 20 | 10 | | |
| 21 | 25 | 2.1739 | **1.5104** |
| 23 | 15 | | |

*4.4. Impact of block-level process to AUC*

In this section, we utilize frame-level AUC to evaluate the effectiveness of block-level process. The results are shown in table 4. The parameters in block-level process are: $h = w = 30$.

**Table 4**
Frame-level AUCs on multiple generation maps and multiple datasets

| Generation Error | | Ped1 | Ped2 | Avenue |
|---|---|---|---|---|
| $GE_{pixel}$ | Frame-level | 0.7966 | 0.8846 | 0.8885 |
| | **Block-level** | **0.8288** | **0.9678** | **0.8982** |
| $GE_{flow}$ | Frame-level | 0.7859 | 0.8694 | 0.8221 |
| | **Block-level** | **0.8666** | **0.98** | **0.8607** |
| $GE_{gdl}$ | Frame-level | 0.7097 | 0.8187 | 0.7855 |
| | **Block-level** | **0.778** | **0.9689** | **0.842** |

As shown in Table 4, the frame-level AUCs are significantly improved after the block-level process in multiple datasets and multiple types of GE maps.

*4.5. Best block-size for block-level process*

In this section, by changing the block-size in block-level process, we prove that the effectiveness of block-level process is not an accidental result, and further, we analyze the best choice to set block-size in each dataset.

Fig. 5 (a) shows the impact of block-size to AUCs on $GE_{flow}$ on multiple datasets. The map-size of $GE_{flow}$ is (384,512). As Fig. 5 (a) shows, with the increase of the block-size, the improvement of block-level process to AUCs first increase and then decrease, all the block sizes are effective to improve the anomaly detection performance, which prove that the effectiveness of block-level process is not accidental. Fig. 5 (b-d) shows the best block-sizes in different datasets. They show that different dataset has different best block-size, and the best block-sizes in all datasets are target-size level. We think the reason for this phenomenon is that anomalies usually caused by targets, the target-size block can contain more information about anomalies, and can reduce the sensitivity to noise.

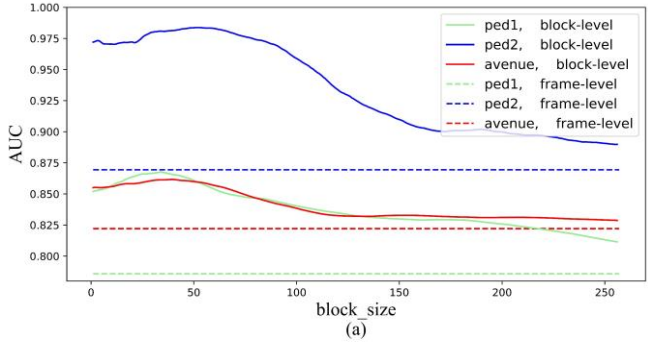

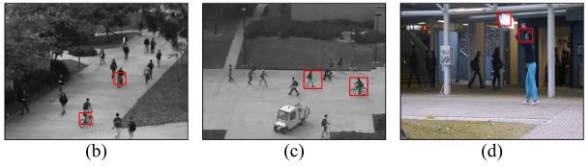

**Fig. 5.** Impact of block-size to anomaly detection. (a) The impact of the block-sizes to AUCs. (b) Best block-size in Ped1. (c) Best block-size in Ped2. (d) Best block-size in Avenue.

*4.6. Impact of different normalize methods*

There are two methods to normalize GEs to calculate anomaly scores: i) Normalize all the GEs in a dataset with uniform standard. ii) Normalize GEs in each video. In this section, we call them norm0 and norm1 respectively.

In this section, we evaluate the impacts of different normalize methods to anomaly detection in two cases: i) We know in advance that all the test videos contain anomalies, we just need to locate the anomaly event. In this case, we use the test samples in the dataset, because every sample in the test set contains anomalies. ii) We do not know whether the test videos contain anomalies, the test videos include normal videos and anomalous videos at the same time. In this case, we use all the videos in the dataset.

Firstly, we visualize the anomaly score curves calculated by these two normalize methods. As shown in Fig. 6, normalizing in whole dataset do not change the shape of the GE curve, while normalizing in each video changes the GE curve a lot and generates strong anomaly scores in every normal video, that will bring false alarms in every normal video segment.

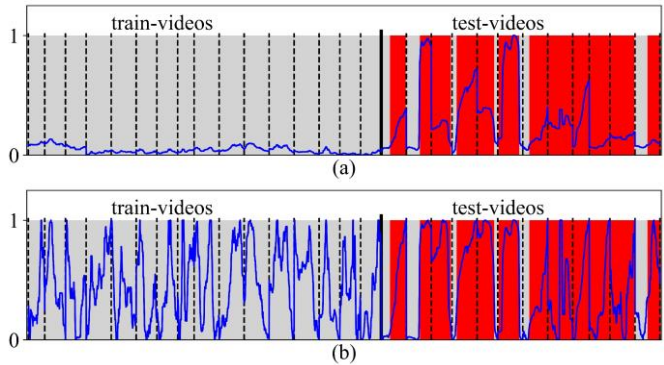

**Fig. 6.** Anomaly scores of Ped2 dataset calculated by different normalize methods. (a) Normalize in whole dataset. (b) Normalize in each video.

Secondly, we evaluate the performance of anomaly scores generated by these two normalize methods on both cases mentioned above. As shown in Table 5, in the first case, norm1 can improve the performances of block-level GEs on some datasets and some GEs, but it's not universal. In the second case, norm1 seriously reduces the anomaly detection performance of the block-level GEs. That prove that, norm1 is not applicable in the case where we do not know whether there exist anomalies in the video.

**Table 5**
The frame-level AUCs at different cases. N0 means Normalize all the GEs in a dataset with uniform standard. N1 means normalize GEs in each video.

| Anomaly Score | | Test data | | | Train + Test data | | |
|---|---|---|---|---|---|---|---|
| | | Ped1 | Ped2 | Avenue | Ped1 | Ped2 | Avenue |
| $GE_{pixel}$ | N0 | 0.829 | 0.968 | **0.898** | **0.940** | 0.976 | **0.938** |
| | N1 | **0.833** | **0.987** | 0.872 | 0.676 | 0.755 | 0.744 |
| $GE_{flow}$ | N0 | **0.867** | **0.98** | **0.861** | **0.953** | **0.998** | **0.896** |
| | N1 | 0.830 | 0.951 | 0.854 | 0.606 | 0.873 | 0.607 |
| $GE_{gdl}$ | N0 | 0.778 | **0.969** | **0.842** | **0.908** | **0.969** | **0.888** |
| | N1 | **0.854** | 0.964 | 0.826 | 0.655 | 0.727 | 0.796 |

In real-world applications, most cases are the second case. Therefore, we adopt norm0 to calculate anomaly scores.

*4.7. Comparison with Existing Methods*

Utilizing equation (9), we compute the final anomaly scores by combining the anomaly scores $s_{pixel}$ generated in $GE_{pixel}$ and anomaly scores $s_{flow}$ generated in $GE_{flow}$: $\hat{s}(t) = s_{pixel}(t) + s_{flow}(t)$. We utilize $\hat{s}(t)$ to detect anomalies. The anomaly detection performance is listed in Table 6, compared with some state-of-the-art approaches. In the block-level process, $h = w = 30$.

**Table 6**
Frame-level AUCs of different methods.

| Methods | Ped1 | Ped2 | Avenue |
|---|---|---|---|
| Conv-AE [4] | 81.1 | 90.0 | 70.2 |
| WTA-AE [27] | 91.9 | 96.6 | 82.1 |
| GNG [30] | **93.75** | 94.09 | N/A |
| Unmasking [44] | 68.4 | 82.2 | 80.6 |
| Conv-LSTM-AE [38] | 75.5 | 88.1 | 77.0 |
| Stack RNN [45] | N/A | 92.21 | 81.71 |
| STAN [5] | 82.1 | 96.5 | 87.2 |
| U-net [6] (baseline) | 83.1 | 95.4 | 85.1 |
| Tang et al. [32] | 84.7 | 96.3 | 85.1 |
| **Our method** | 86.72 | **99.11** | **89.86** |

Compared with the baseline [6], our method improves the frame-level AUC significantly on all the datasets, that proves the effectiveness of the proposed method. Compared with other methods, we achieve state-of-the-art performance on Ped2 and Avenue datasets.

**5. Conclusion**

In this letter, we proposed a promotion method for GE based method: We reduce the interference of the GE of normal regions to anomaly detection by using a block-level process module. Experimentations across multiple datasets show that the proposed method is effective to improve the anomaly detection performance of GE-based method, and achieve state-of-the-art performances on multiple datasets.

**References**

[1] O.P. Popoola, K. Wang, Video-based abnormal human behavior recognition—A review, IEEE Trans. Syst. Man Cybern. Part C Appl. Rev. 42 (2012) 865–878. https://doi.org/10.1109/TSMCC.2011.2178594.

[2] T. Li, H. Chang, M. Wang, B. Ni, R. Hong, S. Yan, Crowded scene analysis: A survey, IEEE Trans.

Circuits Syst. Video Technol. 25 (2015) 367–386. https://doi.org/10.1109/TCSVT.2014.2358029.

[3] B.R. Kiran, D.M. Thomas, R. Parakkal, An overview of deep learning based methods for unsupervised and semi-supervised anomaly detection in videos, J. Imaging. 4 (2018) 36. https://doi.org/10.3390/jimaging4020036.

[4] M. Hasan, J. Choi, J. Neumann, A.K. Roy-Chowdhury, L.S. Davis, Learning Temporal Regularity in Video Sequences, in: Proc. IEEE Conf. Comput. Vis. Pattern Recognit., IEEE, Seattle, WA, 2016: pp. 733–742. https://doi.org/10.1109/CVPR.2016.86.

[5] S. Lee, H.G. Kim, Y.M. Ro, STAN: Spatio-Temporal Adversarial Networks for Abnormal Event Detection, in: 2018 IEEE Int. Conf. Acoust. Speech Signal Process., IEEE, Calgary, Canada, 2018: pp. 1323–1327. https://doi.org/10.1109/ICASSP.2018.8462388.

[6] W. Liu, W. Luo, D. Lian, S. Gao, Future Frame Prediction for Anomaly Detection - A New Baseline, in: Proc. IEEE Conf. Comput. Vis. Pattern Recognit., IEEE, Salt Lake City, UT, 2018: pp. 6536–6545. https://doi.org/10.1109/CVPR.2018.00684.

[7] R. Chalapathy, S. Chawla, Deep learning for anomaly detection: A survey, ArXiv:1901.03407. (2019). https://doi.org/arXiv:1901.03407v2.

[8] Y. Cong, J. Yuan, J. Liu, Sparse reconstruction cost for abnormal event detection, in: Proc. IEEE Comput. Soc. Conf. Comput. Vis. Pattern Recognit., IEEE, Colorado Springs, CO, 2011: pp. 3449–3456. https://doi.org/10.1109/CVPR.2011.5995434.

[9] Y. Cong, J. Yuan, J. Liu, Abnormal event detection in crowded scenes using sparse representation, Pattern Recognit. 46 (2013) 1851–1864. https://doi.org/10.1016/j.patcog.2012.11.021.

[10] B. Zhao, L. Fei-Fei, E.P. Xing, Online detection of unusual events in videos via dynamic sparse coding, in: Comput. Vis. Pattern Recognit., IEEE, Providence, RI, USA, 2011: pp. 3313–3320. https://doi.org/10.1109/CVPR.2011.5995524.

[11] S. Zhou, W. Shen, D. Zeng, Z. Zhang, Unusual event detection in crowded scenes by trajectory analysis, in: 2015 IEEE Int. Conf. Acoust. Speech Signal Process., IEEE, Brisbane, AUSTRALIA, 2015: pp. 1300–1304. https://doi.org/10.1109/ICASSP.2015.7178180.

[12] X. Zhu, J. Liu, J. Wang, C. Li, H. Lu, Sparse representation for robust abnormality detection in crowded scenes, Pattern Recognit. 47 (2014) 1791–1799. https://doi.org/10.1016/j.patcog.2013.11.018.

[13] X. Hu, Y. Huang, X. Gao, L. Luo, Q. Duan, Squirrel-Cage Local Binary Pattern and Its Application in Video Anomaly Detection, IEEE Trans. Inf. Forensics Secur. 14 (2019) 1007–1022. https://doi.org/10.1109/TIFS.2018.2868617.

[14] J. Kim, K. Grauman, Observe locally, infer globally: A space-time MRF for detecting abnormal activities with incremental updates, in: 2009 IEEE Conf. Comput. Vis. Pattern Recognit., IEEE, Miami, FL, USA, 2009: pp. 2921–2928. https://doi.org/10.1109/CVPRW.2009.5206569.

[15] L. Kratz, K. Nishino, Anomaly detection in extremely crowded scenes using spatio-temporal motion pattern models, in: 2009 IEEE Conf. Comput. Vis. Pattern Recognit., IEEE, Miami Beach, FL, 2009: pp. 1446–1453. https://doi.org/10.1109/CVPRW.2009.5206771.

[16] C. Lu, J. Shi, J. Jia, Abnormal Event Detection at 150 FPS in MATLAB, in: Proc. IEEE Int. Conf. Comput. Vis., IEEE, Sydney, NSW, Australia, 2013: pp. 2720–2727. https://doi.org/10.1109/ICCV.2013.338.

[17] V. Mahadevan, W. Li, V. Bhalodia, N. Vasconcelos, Anomaly detection in crowded scenes, in: 2010 IEEE Comput. Soc. Conf. Comput. Vis. Pattern Recognit., IEEE, San Francisco, CA, 2010: pp. 1975–1981. https://doi.org/10.1109/CVPR.2010.5539872.

[18] H. Ren, W. Liu, S.I. Olsen, S. Escalera, T.B. Moeslund, Unsupervised Behavior-Specific Dictionary Learning for Abnormal Event Detection, in: Proc. Br. Mach. Vis. Conf. 2015, British Machine Vision Association, Swansea, UK, 2015: pp. 28.1-28.13. https://doi.org/10.5244/C.29.28.

[19] Weixin Li, V. Mahadevan, N. Vasconcelos, Anomaly Detection and Localization in Crowded Scenes, IEEE Trans. Pattern Anal. Mach. Intell. 36 (2014) 18–32. https://doi.org/10.1109/TPAMI.2013.111.

[20] Y. Yuan, Y. Feng, X. Lu, Structured dictionary learning for abnormal event detection in crowded scenes, Pattern Recognit. 73 (2018) 99–110. https://doi.org/10.1016/j.patcog.2017.08.001.

[21] W. Chu, H. Xue, C. Yao, D. Cai, Sparse Coding Guided Spatiotemporal Feature Learning for Abnormal Event Detection in Large Videos, IEEE Trans. Multimed. 21 (2019) 246–255. https://doi.org/10.1109/TMM.2018.2846411.

[22] R. Hinami, T. Mei, S. Satoh, Joint Detection and Recounting of Abnormal Events by Learning Deep Generic Knowledge, in: Proc. IEEE Int. Conf. Comput. Vis., IEEE, Venice, Italy, 2017: pp. 3619–3627. https://doi.org/10.1109/ICCV.2017.391.

[23] R.T. Ionescu, S. Smeureanu, M. Popescu, B. Alexe, Detecting Abnormal Events in Video Using Narrowed Normality Clusters, in: 2019 IEEE Winter Conf. Appl. Comput. Vis., IEEE, Hawaii, 2019: pp. 1951–1960. https://doi.org/10.1109/WACV.2019.00212.

[24] T.S. Nazare, R.F. de Mello, M.A. Ponti, Are pre-trained CNNs good feature extractors for anomaly detection in surveillance videos?, ArXiv:1811.08495. (2018). https://doi.org/arXiv:1811.08495v1.

[25] M. Sabokrou, M. Fayyaz, M. Fathy, R. Klette, Deep-Cascade: cascading 3D deep neural networks for fast

anomaly detection and localization in crowded scenes, IEEE Trans. Image Process. 26 (2017) 1992–2004. https://doi.org/10.1109/TIP.2017.2670780.

[26] J. Sun, X. Wang, N. Xiong, J. Shao, Learning sparse representation with variational auto-encoder for anomaly detection, IEEE Access. 6 (2018) 33353–33361. https://doi.org/10.1109/ACCESS.2018.2848210.

[27] H. Tran, D. Hogg, Anomaly Detection using a Convolutional Winner-Take-All Autoencoder, in: Proc. Br. Mach. Vis. Conf. 2017, British Machine Vision Association, London, 2017. https://doi.org/10.5244/C.31.139.

[28] D. Xu, Y. Yan, E. Ricci, N. Sebe, Detecting anomalous events in videos by learning deep representations of appearance and motion, Comput. Vis. Image Underst. 156 (2017) 117–127. https://doi.org/10.1016/j.cviu.2016.10.010.

[29] J. Feng, C. Zhang, P. Hao, Online learning with self-organizing maps for anomaly detection in crowd scenes, in: 2010 20th Int. Conf. Pattern Recognit., IEEE, Istanbul, Turkey, 2010: pp. 3599–3602. https://doi.org/10.1109/ICPR.2010.878.

[30] T. Harada, H. Liu, Online growing neural gas for anomaly detection in changing surveillance scenes, Pattern Recognit. 64 (2017) 187–201. https://doi.org/10.1016/j.patcog.2016.09.016.

[31] Y. Fan, G. Wen, D. Li, S. Qiu, M.D. Levine, Video Anomaly Detection and Localization via Gaussian Mixture Fully Convolutional Variational Autoencoder, ArXiv:1805.11223. (2018). http://arxiv.org/abs/1805.11223.

[32] Y. Tang, L. Zhao, S. Zhang, C. Gong, G. Li, J. Yang, Integrating prediction and reconstruction for anomaly detection, Pattern Recognit. Lett. 129 (2020) 123–130. https://doi.org/10.1016/j.patrec.2019.11.024.

[33] M. Ravanbakhsh, E. Sangineto, M. Nabi, N. Sebe, Training Adversarial Discriminators for Cross-Channel Abnormal Event Detection in Crowds, in: 2019 IEEE Winter Conf. Appl. Comput. Vis., IEEE, Waikoloa Village, HI, USA, 2019: pp. 1896–1904. https://doi.org/10.1109/WACV.2019.00206.

[34] M. Sabokrou, M. Khalooei, M. Fathy, E. Adeli, Adversarially Learned One-Class Classifier for Novelty Detection, in: Proc. IEEE Conf. Comput. Vis. Pattern Recognit., IEEE, Salt Lake City, UT, 2018: pp. 3379–3388. https://doi.org/10.1109/CVPR.2018.00356.

[35] Y.S. Chong, Y.H. Tay, Abnormal Event Detection in Videos Using Spatiotemporal Autoencoder, in: ArXiv:1701.01546, 2017: pp. 189–196. https://doi.org/10.1007/978-3-319-59081-3_23.

[36] M. Ravanbakhsh, M. Nabi, E. Sangineto, L. Marcenaro, C. Regazzoni, N. Sebe, Abnormal event detection in videos using generative adversarial nets, in: T. Huang, J. Lv, C. Sun, A. V. Tuzikov (Eds.), 2017 IEEE Int. Conf. Image Process., IEEE, Cham, 2017: pp. 1577–1581. https://doi.org/10.1109/ICIP.2017.8296547.

[37] Y. Zhao, B. Deng, C. Shen, Y. Liu, H. Lu, X.-S. Hua, Spatio-Temporal AutoEncoder for Video Anomaly Detection, in: Proc. 25th ACM Int. Conf. Multimed., ACM Press, New York, New York, USA, 2017: pp. 1933–1941. https://doi.org/10.1145/3123266.3123451.

[38] W. Luo, W. Liu, S. Gao, Remembering history with convolutional LSTM for anomaly detection, in: 2017 IEEE Int. Conf. Multimed. Expo, IEEE, Hong Kong, 2017: pp. 439–444. https://doi.org/10.1109/ICME.2017.8019325.

[39] J.R. Medel, A. Savakis, Anomaly Detection in Video Using Predictive Convolutional Long Short-Term Memory Networks, ArXiv:1612.00390. (2016). http://arxiv.org/abs/1612.00390.

[40] M. Ribeiro, A.E. Lazzaretti, H.S. Lopes, A study of deep convolutional auto-encoders for anomaly detection in videos, Pattern Recognit. Lett. 105 (2018) 13–22. https://doi.org/10.1016/j.patrec.2017.07.016.

[41] M. Sabokrou, M. Fathy, M. Hoseini, Video anomaly detection and localisation based on the sparsity and reconstruction error of auto-encoder, Electron. Lett. 52 (2016) 1122–1124. https://doi.org/10.1049/el.2016.0440.

[42] T. Wang, M. Qiao, Z. Lin, C. Li, H. Snoussi, Z. Liu, C. Choi, Generative Neural Networks for Anomaly Detection in Crowded Scenes, IEEE Trans. Inf. Forensics Secur. 14 (2019) 1390–1399. https://doi.org/10.1109/TIFS.2018.2878538.

[43] M. Sabokrou, M. Pourreza, M. Fayyaz, R. Entezari, M. Fathy, J. Gall, E. Adeli, AVID: adversarial visual irregularity detection, ArXiv:1805.09521. (2018). http://arxiv.org/abs/1805.09521.

[44] R.T. Ionescu, S. Smeureanu, B. Alexe, M. Popescu, Unmasking the abnormal events in video, in: Proc. IEEE Int. Conf. Comput. Vis., IEEE, Venice, Italy, 2017: pp. 2914–2922. https://doi.org/10.1109/ICCV.2017.315.

[45] W. Luo, W. Liu, S. Gao, A revisit of sparse coding based anomaly detection in stacked RNN framework, in: Proc. IEEE Int. Conf. Comput. Vis., IEEE, Venice, Italy, 2017: pp. 341–349. https://doi.org/10.1109/ICCV.2017.45.